%% file: main.tex
\documentclass[sigconf]{aamas}

\usepackage{balance}
\usepackage{latexsym}

\usepackage{amssymb}
\usepackage{amsmath}
\usepackage{amsthm}
\usepackage{booktabs}
\usepackage{enumitem}
\usepackage{graphicx}
\usepackage{color}
\usepackage{algorithm2e}
\usepackage[noend]{algorithmic}
\usepackage{hyperref}
\usepackage{tikz}

\setcopyright{ifaamas}
\acmConference[AAMAS '26]{Proc.\@ of the 25th International Conference
on Autonomous Agents and Multiagent Systems (AAMAS 2026)}{May 25 -- 29, 2026}
{Paphos, Cyprus}{C.~Amato, L.~Dennis, V.~Mascardi, J.~Thangarajah (eds.)}
\copyrightyear{2026}
\acmYear{2026}
\acmDOI{}
\acmPrice{}
\acmISBN{}

\acmSubmissionID{0}

\author{Erel Shtossel}
\affiliation{
  \institution{Bar Ilan University}
  \city{Ramat Gan}
  \country{Israel}}

\author{Gal A. Kaminka}
\affiliation{
  \institution{Bar Ilan University}
  \city{Ramat Gan}
  \country{Israel}
\email{galk@cs.biu.ac.il}}

\newcommand*{\sr}{
	\mathcal{R}
}
\newcommand{\avoid}{
	\mathcal{C}
}
\newcommand*{\prog}{
	\mathcal{P}
}

\newcommand*{\self}{
	\Omega
}
\newcommand*{\diffr}{
	\Delta
}

\title{Modular Reinforcement Learning For Cooperative Swarms}

\begin{abstract}
\input{abstract}

\end{abstract}


\begin{document}
\microtypesetup{expansion=false}
\pagestyle{fancy}
\fancyhead{}
\maketitle


\section{Introduction}
\input{introduction}

\section{Background}\label{sec:background}
\input{background}

\section{Preliminaries}\label{sec:prelim}
\input{preliminaries}

\section{Modular Spatial State Representation}\label{sec:modular}
\input{modular}


\section{Experiments}\label{sec:exp}
\input{experiments}

\section{Conclusions}\label{sec:fin}
\input{fin}

\begin{acks}
This research was supported in part by ISF Grant \# 1373/24.  As always, thanks to K. Ushi.
\end{acks}

\input{main.bbl}

\end{document}

%% file: abstract.tex
A \emph{cooperative robot swarm} is a collective of computationally-limited robots that share a common goal.
Each robot can only interact with a small subset of its peers, without knowing how this affects
the collective utility.
Recent advances in distributed multi-agent reinforcement learning have demonstrated that it is possible for robots
to learn how to interact effectively with others, in a manner that is aligned with the common goal, despite each robot learning independently of others.
However, this requires each robot to represent a potentially combinatorial number of interaction states, challenging the memory capabilities of the robots.
This paper proposes an alternative approach for representing \textit{spatial interaction states} for multi-robot reinforcement learning in swarms. A modular (decomposed) representation is used, where each feature of the state is handled by a separate learning procedure, and the results aggregated.
We demonstrate the efficacy of the approach in numerous experiments with simulated robot swarms carrying out \textit{foraging}.

%% file: introduction.tex
A \emph{cooperative robot swarm} is a collective of robots that work towards a common goal, despite each being limited in the range of its perception and the reach of its actions.
These limitations allow only \emph{local} interactions for each robot, with a small subset of its peers~\cite{correll09,brambilla2013swarm,bayindir2016review,hamann18,dorigo2021a}.

Cooperative swarms pose a significant challenge to swarm robot developers. The maximization of the collective utility by the swarm's members, in aggregate, is encumbered by each individual robot's lack of knowledge about its peers or the state of the collective as a whole.
Often, this is addressed by carefully designing the individual decision-making procedures, such that local decisions are forced to \emph{align} with the collective goal~\cite{rybski98performance,Agression2,song13,ren2023collective}. 
Alternatively, multi-agent reinforcement learning (\textit{MARL}) can be used to learn individual action-selection policies~\cite{kapetanakis02reinforcement,hernandez19survey,kober13,zhang2021multi,kuckling23}, in particular when matched with appropriate aligned rewards~\cite{devlin14,royal25rational}.

A significant challenge to the use of MARL in swarm robots is raised by the limited computational resources of the robots.
Limited memory prohibits explicit representation of the large combinatorial number of states (a \emph{state explosion} problem).
Limited computation capabilities prohibit the use of deep neural networks as function approximations to generalize and
represent states implicitly. Finally, the restrictions on communication range and bandwidth prohibit sharing of information at the swarm level. Thus robots act, for the most part, as \textit{independent learners}, unaware of the decisions of others, nor the individual impact (reward) of these decisions. This is a particularly challenging setting for MARL~\cite{kapetanakis02reinforcement,matignon12,nowe2012game}.

While these challenges are known in general~\cite{lesort18,mohan2023structure,sahni2017statespacedecompositionsubgoal,wong2021state}, the limited computational resources of swarm robots make most solution approaches impractical or even irrelevant.  For perspective, research swarm robots are commonly built around microcontrollers running at a few hundred MHz cycles, using total RAM well under 512KB (the capabilities of popular off-the-shelf robots are presented in Table~\ref{tab:robot_specs}).

\begin{table}[h]
\centering
\resizebox{0.5\textwidth}{!}{%
\begin{tabular}{|l|l|l|l|}
\hline
\textbf{Robot} & \textbf{Controller} & \textbf{Clock Speed} & \textbf{RAM} \\ \hline
\textit{e-puck2}~\cite{epuck2ref} & STM32F4 & 168 MHz & 192 KB \\ \hline
\textit{Elisa-3}~\cite{elisa3ref} & ATmega2560 & 8 MHz & 8 KB \\ \hline
\textit{Pololu3Pi}~\cite{pololu3pi} & ATmega328 & 20 MHz & 2 KB \\ \hline
\textit{Krembot}~\cite{photonref} & Particle Photon (ARM Cortex M3) & 120 MHz & 128 KB \\ \hline
\textit{Arduino Robot}~\cite{arduinobot} & Two ATmega32u4 & 16 MHz & 2.5 KB each \\ \hline
\textit{Kilobots}~\cite{kilobots} & ATmega328 & 8 MHz & 32 KB \\ \hline
\end{tabular}%
}
\caption{Specifications of popular off-the-shelf robots.}
\label{tab:robot_specs}
\end{table}

Multi-robot swarm foraging, a canonical cooperative swarm task, is a good example of how the severe computational limitations of swarm robots pose a challenge to reinforcement learning approaches.
In foraging, multiple robots search a shared arena for items that they pick up and bring to their base. The performance of the swarm is measured by the sum total of collected items.
Robots have limited sensor and communication ranges, and cannot communicate with all other robots; in some cases, they cannot communicate at all.  To navigate, each robot typically represents its spatial interaction state, composed of the range and bearing to all robots and obstacles in its sensed surroundings. As long as only the current state is maintained, this is not a challenge. However, if a naive reinforcement learning mechanism needs to learn how to act in each state, then all possible spatial states must be represented. The number of states grows combinatorially large with the number of entities and the resolution of measurements.  A latent representation using deep neural networks is not feasible given the limited computation available (some controllers do not support floating-point operations in hardware).

To address the state explosion problem, this paper presents a modular (decomposed) state representation, where each state feature (e.g., the position of each neighbour) is represented separately and handled by a separate learning process.
In this way, rather than a single learner addressing a $k$-factor state representation ($\mathcal{O}(2^k)$ states to be stored),
there are $k$ simple learners, requiring storing a total of $\mathcal{O}(k)$ states. The learners' recommendations are aggregated via a fixed action-fusion mechanism.

We describe the modular representation in detail and evaluate its use in a comprehensive set of experiments with simulated robots in a cooperative foraging task. The experiments show that the modular representation works on par with a full state learning algorithm, despite its modest memory and computational requirements.

%% file: background.tex
Research on robot swarms focuses on the collective behaviour of decentralized, self-organized systems composed of numerous computationally-limited robots. The research area is very broad, covering many different tasks and approaches.
We refer the reader to comprehensive surveys of the area~\cite{correll09,brambilla2013swarm,bayindir2016review,hamann18,dorigo2021a}. 

We focus on \textbf{cooperative swarm foraging}, an operation where a group of swarm robots is tasked with continually searching for objects of interest (\emph{items}), and when found, transporting them to collection points (\emph{bases}). This is a canonical swarm task with many studied variants:
see
\cite{shell2006foraging,winfield09,foraging-survey17,lu2020swarm,ren2023collective} for comprehensive surveys.
%

This paper focuses on using \emph{multi-agent reinforcement learning} (MARL)~\cite{kapetanakis02reinforcement,yang2004multiagent, Busoniu2008,kober13,hernandez19survey,zhang2021multi,kuckling23,fatima24} for swarm foraging.
MARL is particularly challenging in swarm settings, because each agent is an \emph{independent learner}~\cite{panait2003collaborative,icra10dan,matignon12,nowe2012game,marden13,marden18}, 
unable to get feedback on the effects of its individual action on its nearby peers, nor on the swarm as a whole.  Nevertheless, a recent line of work has proposed methods for \emph{aligning} the individual rewards received by each robot, such that their maximization by the individual necessarily optimizes the swarm goals. In particular, \emph{difference rewards}~\cite{devlin14}, 
which computes the marginal contribution of each robot to the swarm, offers a promising approach to such alignment. By converting robot collision overheads to internal measurement of time, it is possible for swarm robots to approximate the difference reward, in a completely distributed manner, without relying on any communication or public signals~\cite{aamas19,frontiers25,royal25rational}. This is the approach we take here.

\paragraph{State Representation.}
When reinforcement learning is applied to swarm robots, a critical challenge emerges in choosing the state representation to match the limited memory and computation resources available. The problem of \textit{state-explosion}---the combinatorial growth in the number of states as necessary state features are represented---is, of course, well known. It can be tackled in several ways: state \textit{abstraction} \cite{mohan2023structure,lesort18} or state \textit{decomposition} \cite{mohan2023structure,sahni2017statespacedecompositionsubgoal,wong2021state}.


State \textit{abstraction} removes state features or replaces them with more general features that summarize important state attributes, while reducing the number of features.
Some attempt to learn abstractions~\cite{kokel21} or dynamically choose between them~\cite{maql}. Fully abstracting the state leads to a state representation that includes a single state (sometimes called \emph{stateless}), reducing the stateful reinforcement learning problem to a multiarm bandit. This can be effective~\cite{icra10dan,aamas19}, but tends to be brittle~\cite{aamas19}. 

Often, domain experts design state-abstractions manually. For instance, and specifically for the use of learning in foraging, Douchan et al.~\cite{aamas19} abstracted collision states in foraging by measuring the local density (i.e., the number of neighbours). The loss of detail can be detrimental to the learning task. For instance, Fig. \ref{fig:MultiStateBlast} shows two states from the perspective of the marked focal robot. These differ in their \textit{geometrical detail}, but share the same \emph{local density}. The safe actions in the two states are different, but relying on the density information as proposed in~\cite{aamas19} would make the two states appear identical.

\input{figures/state_figure.tex}

In contrast to abstraction, state \textit{decomposition} breaks the state into smaller sub-states that can be input into the same single RL algorithm to return an action per sub-state~\cite{LOPPENBERG2024102812,mohan2023structure}. The actions from different sub-states are then aggregated into a single chosen action for the agent. This results in a state representation that grows linearly. At the extreme, a \emph{modular representation}, whereby each different state feature is addressed by separate learning processes, reduces the number of states to be linear in the number of features: for $k$ state features, we utilize $k$ processes, each handling $n$ possible states of the specific feature assigned to it~\cite{mohan2023structure}.


In this paper, we develop and evaluate a modular state representation for spatial states. Spatial states are very commonly used in robots (sometimes they are known as local obstacle maps, or local occupancy grids), as they form the backbone for navigation in the presence of objects and other robots. A naive (fully factored) state representation of such states tends to explode quickly. For instance, in Fig.~\ref{fig:MultiStateBlast}, the focal robot is positioned in the center of a 5-by-5 discrete sensing grid. Simply marking 0 or 1 to note the presence of a robot or object in the 24 possible locations around it, yields $2^{24}=16777216$ states. In contrast, a modular representation for the same type would have $2\cdot 24=48$ states in total, divided among 24 learning processes.

%
%

%% file: figures/state_figure.tex
\tikzset
{%
  pics/matrix/.style n args={6}{
    code={%
      \begin{scope}[y=-1cm, scale=0.6]
        \draw    (0,0) grid (#2,#1);
        \node at (0.5*#2,-0.5)   {#3};
        \node at (0.5*#2,#1+0.5) {#4};
        \node at (-0.5,0.5*#1)   {#5};
        \node at (#2+0.5,0.5*#1) {#6};
      \end{scope}
    }},
}

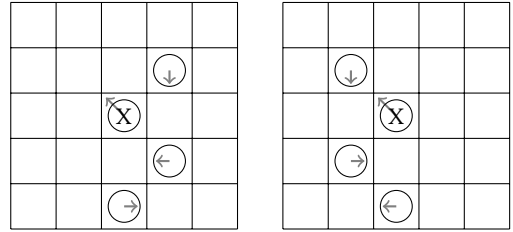
\begin{figure}[htbp]
\begin{tikzpicture}
  \begin{scope}[scale=0.6]


    \pic at (-1,-1)   {matrix={5}{5}{}{}          {}{}};
    \pic at (5,-1)   {matrix={5}{5}{}{}          {}{}};

    \coordinate (A) at (1.5,-3.5);
    \coordinate (B) at (1.5,-5.5);
    \coordinate (C) at (2.5,-4.5);
    \coordinate (D) at (2.5,-2.5);
    \coordinate (E) at (7.5,-3.5);
    \coordinate (F) at (7.5,-5.5);
    \coordinate (G) at (6.5,-4.5);
    \coordinate (H) at (6.5,-2.5);

    \draw[->, thick, gray] (A) -- (1.1,-3.1); 
    \draw[->, thick, gray] (B) -- (1.8,-5.5); 
    \draw[->, thick, gray] (C) -- (2.2,-4.5); 
    \draw[->, thick, gray] (D) -- (2.5,-2.8); 
    \draw[->, thick, gray] (E) -- (7.1,-3.1); 
    \draw[->, thick, gray] (F) -- (7.2,-5.5); 
    \draw[->, thick, gray] (G) -- (6.8,-4.5); 
    \draw[->, thick, gray] (H) -- (6.5,-2.8); 

    \draw (A) circle (10pt) node {X};
    \draw (B) circle (10pt) node { };
    \draw (C) circle (10pt) node { };
    \draw (D) circle (10pt) node { };
    \draw (E) circle (10pt) node {X};
    \draw (F) circle (10pt) node { };
    \draw (G) circle (10pt) node { };
    \draw (H) circle (10pt) node { };

  \end{scope}

\end{tikzpicture}
\caption{Two states that have the same density, and mirrored geometry. The same action is safe in one, dangerous in the other.}
\label{fig:MultiStateBlast}
\vspace{12pt}
\end{figure}

%% file: preliminaries.tex
We begin by describing the foraging swarm task and draw connections to Markov games so as to pose the individual robot's action-select mechanism as a reinforcement learner, engaging in a highly-distributed multi-agent reinforcement learning process.

\paragraph{Learning Opportunities in Swarm Foraging}\label{sec:foraging}
From the perspective of a single swarm robot, the foraging task can be divided into several distinct operational states. A straightforward design separates \textit{item-related operations} (searching for items, picking them up, navigating back to base, dropping the items) from \textit{social operations} (monitoring for potential collisions, selecting collision-avoidance and navigation actions, managing communications if available). There are opportunities for learning---and in particular for reinforcement learning---in many of these operational states. However, we focus on a specific operation: responding to a collision if one is imminent, i.e., \textit{collision-avoidance}.


Collision-avoidance involves significant use of spatial state representation. In non-learning collision-avoidance procedures, the current local state is used as the backbone to any computation of a response, whether myopic or predictive. However, as such procedures do not learn, they only use a single represented state at a time (the current state), and so do not reach the memory limits of simple robots. Such spatial state maps include information about the range and bearing of every robot within the sensing radius; they can sometimes include information about the heading of each such robot, or other information of value. Learning-based methods, however, must store multiple states to respond appropriately to each.

From the point of view of a collision-avoidance learning method, the action-selection point occurs with every collision, where a collision-avoidance action is selected. The reward is received later, as an evaluation of the success of the action in resolving the collision state. In other words, the collision-avoidance learning process discussed in this paper experiences foraging as series of collision-avoidance action-selection opportunities: every impending collision triggers the learner to select an action, and the reward is received upon evaluation---which may be immediate or may be given only on reaching the next collision state (as is commonly modeled in reinforcement learning, the reward $R(s,a,s')$ is given for an action $a$ selected in state $s$, and reaching the new state $s'$).


\paragraph{Foraging as a Markov Team Game}
\label{sec:markov}
We adapt the perspective of~\cite{royal25rational} where the foraging task is modeled as an infinite-horizon, fully-cooperative Markov game (i.e., a Markov Team Game with infinite horizon), where all $N$ members of the swarm interact when colliding. The game is composed of an endless (unknown termination) sequence of \emph{stage} games, where each stage is defined by a collision event as described above. For brevity, we describe below only the necessary formal components needed to define the reinforcement learning problem and methods stemming from this model. We refer the reader to~\cite{royal25rational} for details.

The collective reward of the swarm is modeled as follows, as the undiscounted \textit{average collective rewards} of a joint policy $\pi$:
\begin{align}
	\sr(\pi) := &  \lim_{K\rightarrow\infty}\frac{1}{K} \sum_{k=1}^K \sum_{{s_k}\in S}\left[  R_{k}(\pi)\cdot D_{k}(\pi) \right]
	\label{eq:mean-sr}
\end{align}

Here, $K$ is the number of stage games played (i.e., the horizon tending towards infinity), $R_k(\pi)$ is shorthand notation for $R_k(s_{k-1},\pi(s_{k-1}),s_{k})$, the \emph{joint} reward received for taking an action according to the policy, in stage $s_{k-1}$ resulting in reaching state $s_k$. Similarly, $D_k(\pi)$ is shorthand for the transition probability function $D_k(s_{k-1},\pi(s_{k-1}),s_{k})$, i.e., the probability of reaching $s_k$ given the application of the policy. Note that all terms here are \emph{joint}: joint states $s_{k-1}, s_k$ (the combination of all agents' states), a joint action $a$ (the combination of all individual actions taken by the agents), and so forth.

Fully-cooperative (team) games have the property that the payoff of any given joint action (the combined individual action of all agents) is \emph{shared} by all, in the sense that all agents get the same reward. In this purely theoretical perspective, if all the agents know $\sr$, the collective reward, they can directly measure the effects of the individual actions. Under these settings,  it is straightforward to show (i) that their individual self-interested selections will lead to a Nash equilibrium (i.e., the strategy guarantees stability), and (ii) that the  Nash equilibrium also maximize $\sr$ (i.e., it guarantees collective reward optimality). Allowing the agents to be rational and self-interested in maximizing their rewards will necessarily maximize the collective rewards, as they are one and the same.

Here, the rewards of the collective are said to be \emph{aligned} with those of the individual~\cite{wolpert02,devlin14}. In this case, this simply is a result of all agents sharing the rewards, so alignment is trivial. Using reinforcement learning, the agents may evaluate different strategies $\pi$ by the collective rewards $\sr(\pi)$, until converging to the optimal strategy $\pi^*$. 

\paragraph{Individual, distributed learning.}\label{sec:dl}
The model (Eq.~\ref{eq:mean-sr}) is a purely theoretical abstraction of actual swarm tasks in general, and of swarm foraging in particular. In reality, no instantaneous reward can be shared; The collective reward $\sr$ is unknown to the robots, who only know their own actions, not those of their peers. This is the challenging case of \emph{independent learners}~\cite{matignon12}. Thus, instead, the $N$ robots are faced with the collective reward definition based on their individual rewards $r^i$, where $1\leq i\leq N$ denotes an agent.

\begin{align}
	\sr(\pi) := &  \lim_{K\rightarrow\infty}\frac{1}{K} \sum_{k=1}^K \sum_{i=1}^{N} \sum_{{s_k}\in S} \left[ r^i_{k}(\pi^i)\cdot d^i_{k}(\pi^i) \right]
	\label{eq:individually-sr}
\end{align}

\paragraph{Individual rewards.}\label{sec:dr}
Through the years, different individual rewards have been proposed for the use of the functions $r^i(\cdot)$. A promising approach was introduced in a series of papers~\cite{icra10dan,aamas19,frontiers25,royal25rational}, which seeks to translate the unknowable utility (e.g., number of items collected) into a measure accessible to individual robots, by assuming that the length of time in which robots were active in \emph{item-related tasks} (as discussed above) provide gains for the swarm, while \emph{social tasks}, in particular handling collisions, is detrimental and should be avoided.

For each collision stage $k$, the robot selects an individual collision-avoidance action $a_k$. It then measures the duration of the interval until the collision is resolved, and the duration of the subsequent interval in which the robot was free to engage in productive foraging activities (searching for items, collecting them, etc.). The duration of the collision-avoidance interval for agent $i\in N$ is denoted  $\avoid^i_{k}(\pi^i)=\avoid^i(s_{t_{k-1}},\pi^i(s_{t_{k-1}}),s_{t_{k}})$, and the duration of the subsequent productive foraging interval is denoted $\prog^i_{k}(\pi^i)=\prog^i(s_{t_{k-1}},\pi^i(s_{t_{k-1}}),s_{t_{k}})$. Their sum is the total duration of the stage $\left[\avoid^i_{k}(\pi^i)+\prog^i_{k}(\pi^i)\right]$.

Various reward functions based on this approach have been proposed, including individual self-interest rewards like stage overhead
\begin{align*}
	EI(\cdot) := \frac{\avoid^i_{k}(\pi^i)}{\left[\avoid^i_{k}(\pi^i)+\prog^i_{k}(\pi^i)\right]},
\end{align*} introduced in~\cite{icra10dan},
or its simpler variant, self-interest reward
\begin{align}
\self := \prog^i_{k}(\pi^i)-\avoid^i_{k}(\pi^i),
\label{eq:self}
\end{align}
which we will utilize in some of the experiments. These self-interested rewards are generally not aligned with the swarm collective. Their difference-reward~\cite{devlin14} variants do carry such a guarantee: $AEI$ is the aligned version of $EI$, presented in~\cite{aamas19,frontiers25}. $\diffr$, defined in~\cite{royal25rational} is what we use in this paper.
\begin{align}
		\Delta_{k}^i(\pi) :=
		 \beta^i\prog_{k}^i(\pi^i)  - \alpha^i\avoid_{k}^i(\pi^i) -(n-1)(\alpha^i+\beta^i)\overline{\avoid_{k}^i}(\pi^i),
		 \label{eq:diffr}
\end{align}
where $\alpha, \beta$ are individual constants defining the rate of productivity loss in collisions, and the rate of productivity gain when foraging, resp. We abuse the notation of $N$ slightly, using it here to denote the set of agents, rather than their number, so $i,j\in N$ denotes agents.

%% file: modular.tex
We propose to decompose the full spatial state of the robot by its features, creating a modular representation whereby each state feature is handled by an independent learning process (described in Section~\ref{sec:mod-learner}), and the recommendations of all the learners are aggregated by a fixed procedure (nicknamed, the \emph{council}), which makes a decision on the final action to be taken (described in Section~\ref{sec:mod-council}).

\subsection{A Single Learner per Spatial Feature}
\label{sec:mod-learner}
We divide up the spatial state features by direction. Specifically, for a set of sensors facing a particular direction, we assign a separate learning process. The identification of a particular process with a fixed direction means that the direction feature is implicit in the process and does not need to be explicitly represented.

The various states of each process are composed of the values the sensor(s) can take (jointly).  Thus for example a range sensor facing in a particular angle relative to the robot heading (e.g., 45 degrees to the left of the forward direction) is going to take on values related to range: binary values (0/1) can be used to simply indicate the presence of a neighbouring robot within collision radius, while a greater range can indicate other measurements (up to the distance measured).

The learning process must select an action in response to the sensed state. These actions are selected from the action space of \emph{the entire robot}, and are not restricted to actions associated with the direction represented by the learning process. This freedom enables sensors to recommend actions that drive the robot not only in the opposite direction, but also in other approximate forward directions.

We note a subtle but critical caveat with respect to the choice of action-spaces: the actions must be \emph{composable}, as the action for the robot will be decided by a fusion mechanism, fusing together the actions preferred by the modular learning processes, to generate a single action used by the robot. In particular, this presents a departure from the previous work discussed above~\cite{aij07avi,icra10dan,aamas19,frontiers25,royal25rational}, all of which used \emph{collision-avoidance algorithms} as the robot's actions. These serve as macro-actions, but cannot be composed and merged arbitrarily, and perhaps even not at all. In the experiments, we will evaluate the use of algorithmic actions vs composable actions (direction vectors, constant speed).

Given the state-space and action-space as defined above, a reinforcement learning algorithm may now be used to drive the learning process. Its task is to learn a policy mapping the sub-states (values of the associated spatial feature), with recommended actions for the robot.
It is important that all the learning processes are triggered together whenever the robot needs to select an action. It can be tempting to think of modular learning as an asynchronous process (no collision in this direction? No problem!), but this is fundamentally incorrect. Breaking down the state into modular features means that the learning occurs in a synchronous manner: all processes are engaged in selecting actions. A corollary of this is that the reward received by selecting the fused action is shared among all the independent learners. There is no attempt at credit assignment.

The implementation used in the experiments serves to ground and clarify the presentation above. The experiments utilize cylindrical robots with 8 range sensors uniformly distributed around their circumference. We therefore have 8 sensing directions, and thus 8 spatial features to model. Each is assigned an identical learning process.  We chose to use binary range values, so there are only two states, corresponding to the detected presence of a neighbour within collision distance (\emph{collision}, or lack thereof (\emph{clear}).

The action-space used in the experiments (termed \emph{vectorial}) is made of direction vectors, in the 8 directions used by the sensors. This is for simplicity---there is no immediate connection between the state-space and the action-space, other than the restriction on using composable actions. The robot speed is fixed, and the actions choose direction.

The learning process is triggered if any sensor detects a neighbour within the collision radius. Then \emph{all} modular learning processes are triggered.
Every process responds independently as follows: if the state is \emph{clear}, it responds with a fixed action (recommending its own direction). Otherwise, it uses a multi-armed bandit learning algorithm (Continuous UCB-1~\cite{Eden}) to select one of the eight direction actions. The reward used in most experiments is the difference reward ($\diffr$, Eq.~\ref{eq:diffr}), as it is aligned, and has been demonstrated successfully in non-modular learning for foraging.

\subsection{The Council}
\label{sec:mod-council}

Each of the independent learners selects an action, treated as a preference for the action to be taken by the robot. The aggregation procedure, nicknamed \emph{council}, is used to fuse the preferences to generate compromises.

The learners' preferred vectors disagree (e.g., half the learners prefer one direction, the other half its opposite) or simply offer no clear compromise (e.g., each of the sensors recommends a different direction, so no direction is preferred by more than one learner). This is reminiscent of behavior-fusion control in robotics~\cite{rosenblatt89}.

We use the following fixed procedure (Alg.~\ref{alg:councilweight} below) for generating a selected action that the robot will execute. It accepts the preferred action of each process $A_i$, and uses it to give a high preference value to the direction indicated. It also gives a (lower) preferred value, based on a Gaussian distribution, to neighbouring directions, to promote satisfying actions~\cite{rosenblatt89}.
Then the preferences are accumulated, and the accumulated distribution is normalized to generate a discrete probability distribution over the actions. Finally, an action is selected by sampling the probability distribution. The council process then monitors the execution of the agent, and passes the reward to all independent learners---with no credit assignment.

\begin{algorithm}[htbp]
\caption{Directional Probability Distribution via Double Gaussian}
\label{alg:councilweight}
\LinesNumbered

\SetKw{KwDefine}{define}
\KwIn{Direction angles $D \in \{0, 45,90,135,180,225,270,315\}$, standard deviation $\sigma = 2$}
\KwOut{Normalized probability vector $P \in \mathbb{R}^8$}

\KwDefine{$A \gets \{0^\circ, 45^\circ, 90^\circ, 135^\circ, 180^\circ, 225^\circ, 270^\circ, 315^\circ\}$}\\
\KwDefine{$P \gets \{0, 0, \dots, 0\}$}\\
\ForEach{$d \in D$}{
	\For{$i \gets 1$ \KwTo $8$}{
		$\text{diff} \gets |d - A_i|$\\
		\If{$\text{diff} > 180$}{
			$\text{diff} \gets 360 - \text{diff}$\\
		}
		$P[i] \gets P[i] + \exp\left(-\frac{1}{2} \left( \frac{\text{diff}}{\sigma} \right)^2 \right)$\\
	}
}
Normalize $P$ such that $\sum_{i=1}^{8} P[i] = 1$\\
\Return $P$

\end{algorithm}

%% file: experiments.tex
We evaluate the modular representation in high-fidelity physical simulation, rather than physical robots. This is to allow long training times as needed, and to be able to contrast with algorithms whose computational requirements (run-time and memory) may not be suitable for physical robot resources.  

We first describe the experiment settings in Section~\ref{sec:setup}. In Section~\ref{sec:headtohead} we report on the main set of results, contrasting the performance of foraging in policies learned with the modular representation, with algorithms using full state representations, and fixed navigation. We then present results from evaluating the robustness of the performance to reward changes (Section~\ref{sec:robust}) and show that the move to the simpler, vectorial action space, necessitated by the modular representation, does not, in general, hurt performance (Section~\ref{sec:algspace}).

\subsection{Experiment setup}\label{sec:setup}
We use the ARGoS3 robot swarm simulator~\cite{argos} to simulate Krembot robots.
These are prototypical research-grade swarm robots. They are approximately cylindrical: the diameter is 6.5cm, and the height is 10.6cm. They have 8 visual and range sensors spread uniformly around the robot's circumference, allowing the distinction of color and distance in 8 directions.


We have constructed three simulated environments of 1.5 square meters with a home base of 0.3-meter diameter and randomly scattered 0.1-meter diameter pucks. When a robot picks up a puck, a new puck will spawn around the arena in a uniform distribution.
\begin{itemize}
    \item \textbf{Arena 1}: The home base is in the middle of the arena
    \item \textbf{Arena 2}: The home base is in the corner of the arena
    \item \textbf{Arena 3}: There are two home bases, close to opposite corners.
\end{itemize}

\begin{figure}[h]
    \centering
    \begin{minipage}[t]{0.15\textwidth}
        \centering
        \includegraphics[width=\linewidth]{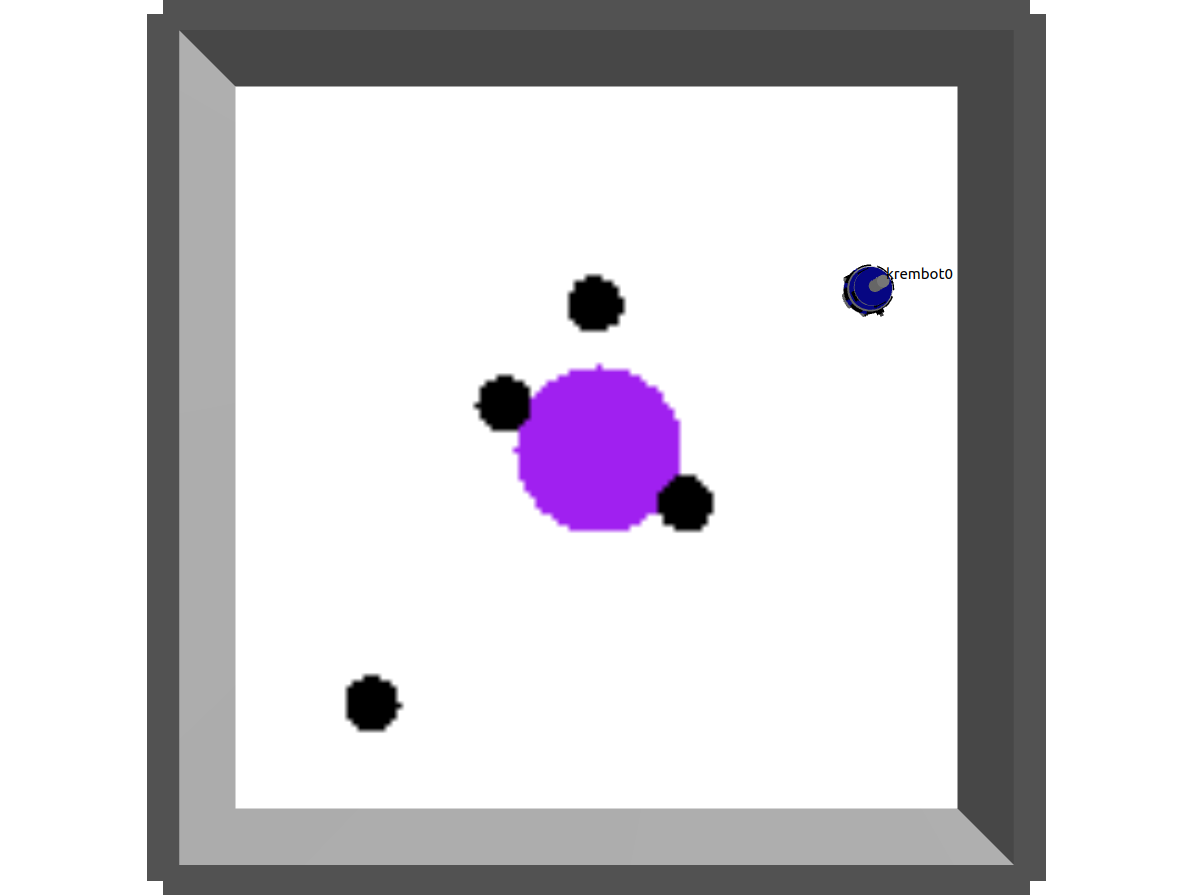}
    \end{minipage}
    \hfill
    \begin{minipage}[t]{0.15\textwidth}
        \centering
        \includegraphics[width=\linewidth]{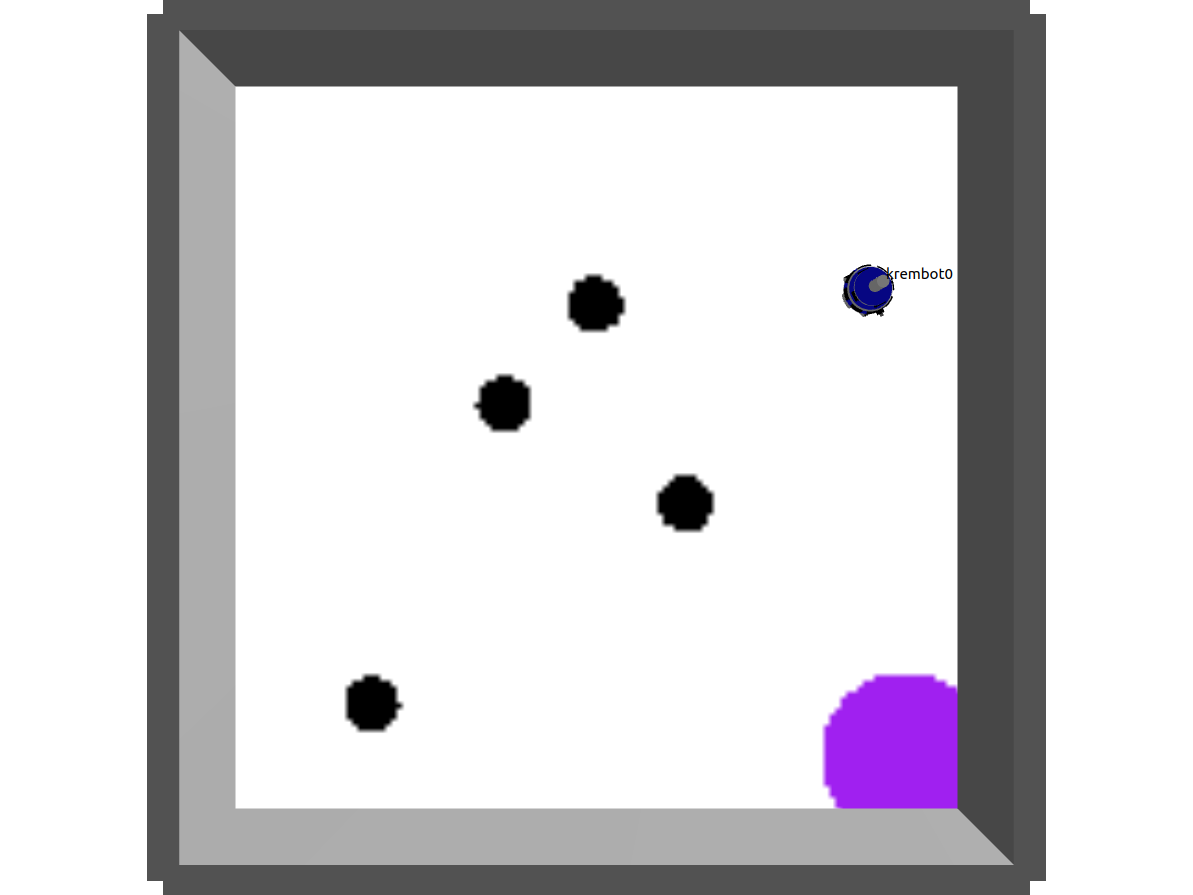}
    \end{minipage}
    \hfill
    \begin{minipage}[t]{0.15\textwidth}
        \centering
        \includegraphics[width=\linewidth]{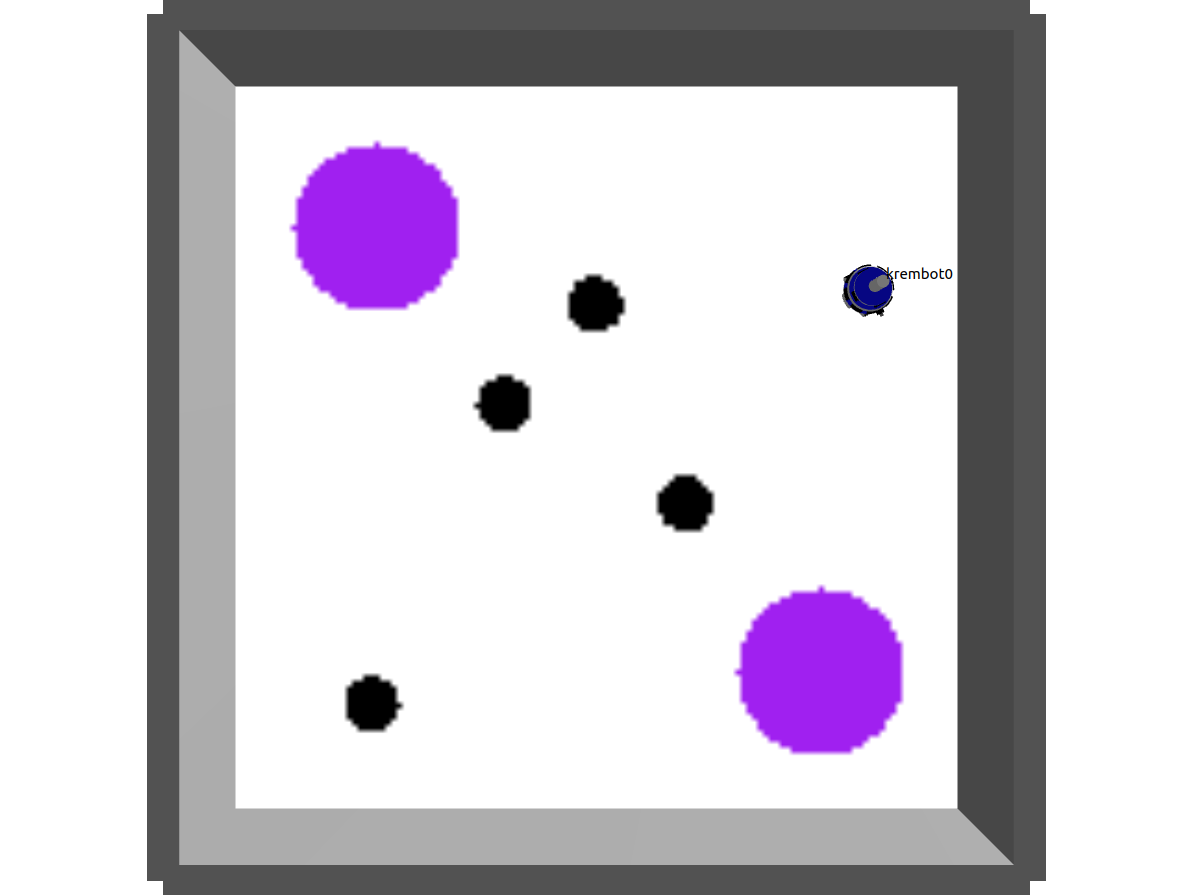}
    \end{minipage}

    \caption{Arenas 1 (left), 2 (middle), and 3 (right). Purple areas show the home base, the black dots are the pucks, and the blue dot is a Krembot for size reference.}
\end{figure}

The number of robots was varied (4--36 in jumps of 4). 
Each test was run with twenty different seeds. These seeds determine the random placement of items in the arena and the initial random positions of the robots.

During the foraging, every robot followed the same basic algorithm: When no robots are detected in the \textit{collision radius} of four centimetres around the robot, the robot wanders (random walk) in search of pucks. If it gets to the collision radius from a wall, it will turn towards the arena at a random angle. When a puck is found, the robot is informed where the closest home base is located and its own location, then the robot will calculate the direction to the base and navigate back to it. If at any point a robot is detected in the collision radius, it will stop foraging and call the avoidance method to solve the conflict. Pucks do not fall upon collision, they can only be dropped intentionally, at the base.

\subsection{Modular representation vs. full-state RL}\label{sec:headtohead}
We begin by evaluating the performance of the modular representation as described in Section~\ref{sec:modular}. First, we instantiate the modular method for the simulated robots. Each sensor is assigned a separate multi-arm bandit learner: here, a continuous-time UCB1 algorithm~\cite{Eden}, with a learning rate of $0.1$. It chooses between one of 8 directions; a total of 8 learners make these recommendations, and the council aggregates them. We compare the performance with three alternatives:

\begin{itemize}
    \item \textbf{Random}, which selects actions randomly. This serves as a baseline; there is no representation of the state at all.
    \item \textbf{Dynamic Window}~\cite{dynamic-window97}, a fixed collision-avoidance procedure. The robot will go to the free space that is closest to its own current direction. This algorithm is \textit{stateful} because it knows where other robots are and what their own direction is.
    \item \textbf{R-Learner}, a myopic version of the algorithm presented in \cite{schwartz93} with learning rate and exploration rate of $0.1$. This algorithm is \textit{stateful}, and would be too expensive to run on a physical robot. It therefore serves as an upper-bound for how good a full reinforcement learning algorithm may perform. Still, states are represented by assigning a boolean value per sensor, corresponding to whether a neighbour is sensed within the \textit{collision radius}. As there are 8 sensors, the total number of states is 256.
    \item \textbf{Continuous-Time Q-Learning}, an adaptation for Q-Learning to accommodate for differences in the amount of time that passed at every step. The learning rate for step t $\alpha_t$ is determined by: $\alpha_t = 1-e^{-\lambda\tau}$, where $\tau$ is the duration of the stage.
\end{itemize}

We chose to use a simple Boolean state representation to compute and compare the Continuous-Time Q-Learning and R-Learning: in each of eight directions where a range sensor is pointing, the value is 0 (false) if no neighbour is sensed, or 1 (true) if a neighbour is sensed. The total number of states in this representation is then $2^8=256$ for the full-state learners, and $2\times 8=16$ for the modular representation.

Adding more information can be beneficial to all of the learners---including the modular method---but increases the state space combinatorially (for the full-state learners) or multiplicatively (modular method). For example, the value of each of the eight direction features can take on 0 (no neighbour is sensed), or 1--8 (the heading of the sensed robot, in terms of one of the eight directions), for a total of 9 values. Certainly, knowing whether a sensed robot is heading away from the observer or towards it is of great value.  In this case, the state space will expand to $9^8=43,046,721$ states. In contrast, the modular method will use $9\times 8=72$ states.

Thus, the use of boolean features is consistent with our goal of evaluating the full-state learners and the modular learner under practical limitations closer to the reality of physical swarm robots. In addition, a greater number of states easily extend the learning period, a common phenomenon in reinforcement learning.

All learning algorithms were allowed to learn for twelve simulated hours. After the learning stage, the learning is stopped, and the evaluation stage of twenty simulated minutes starts. Algorithms without learning are also evaluated for twenty minutes.

\begin{figure}[htbp]
    \includegraphics[width=0.8\linewidth]{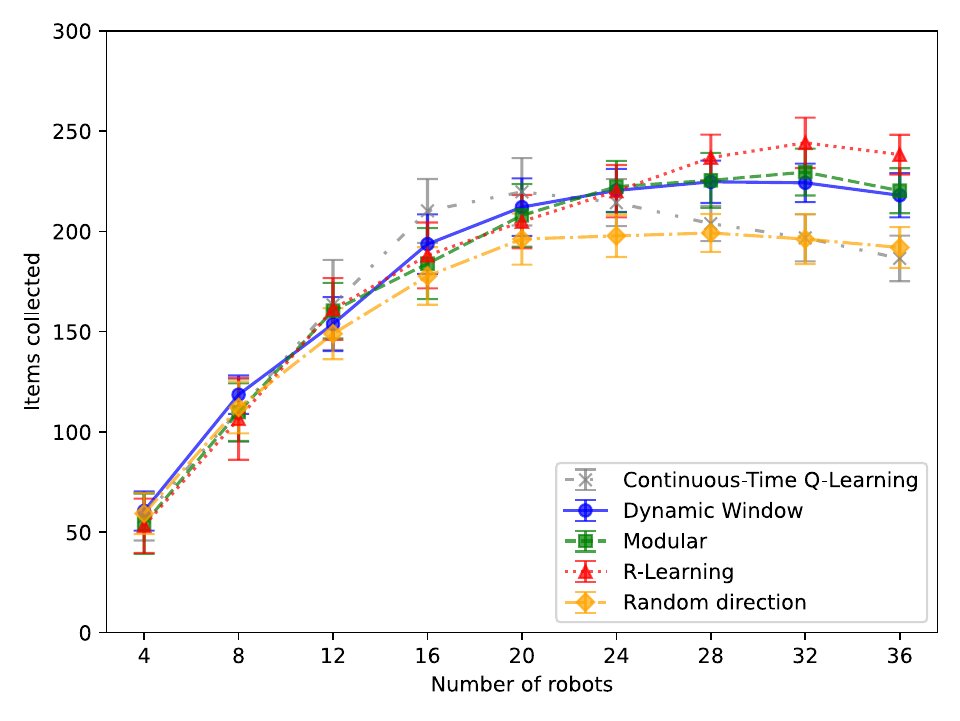}
    \caption{Results in Arena 1}\label{fig:mod1}
\end{figure}

Figure~\ref{fig:mod1} shows the results. The Y-axis measures the number of items collected during the evaluation phase of the experiments, while the X-axis shows the number of robots. Each line shows the results for a different algorithm. The points mark mean results over 20 runs (different seeds), while the error bars mark standard errors.

We observe in Figure~\ref{fig:mod1} that up to twelve robots, all algorithms provide essentially indistinguishable results. Dynamic window performs well in lower-density scenarios. At 28 robots the learning algorithms gain an advantage: the stateful R-Learner performed better once the density of the robots is sufficiently high, so that the more informed (and more expensive to store) representation gives an advantage. The modular approach performs stably, on par with dynamic window.

\begin{figure}[htbp]
    \includegraphics[width=0.8\linewidth]{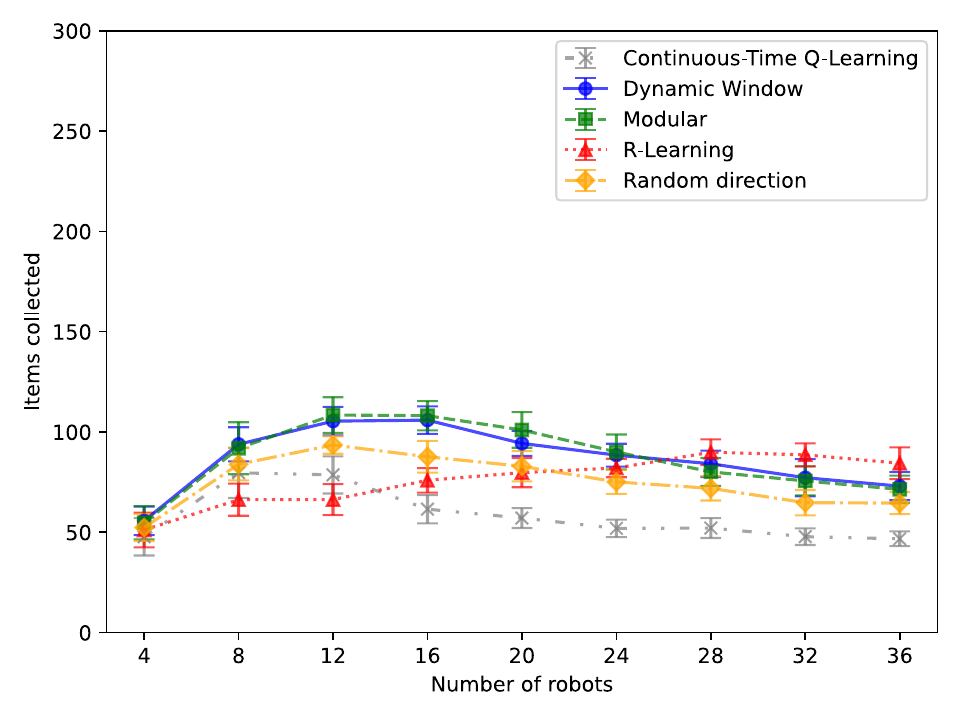}
    \caption{Results in Arena 2}
    \label{fig:mod2}
\end{figure}

Figure~\ref{fig:mod2} shows the results from Arena 2. Here, the base is close to the corner, resulting in a harder task as more collisions occur when the robots try to return the pucks. All methods are affected by the challenge of this task, performing up to $2.5$ worse in comparison to Arena 1 (Fig.~\ref{fig:mod1}). Here, too, modular learning closely matches dynamic window, better than R-learning in lower densities, but falling as the density rises. R-learning seems to work only when the density is very high, even losing to the random algorithm in low densities.

\begin{figure}[htbp]
    \includegraphics[width=0.8\linewidth]{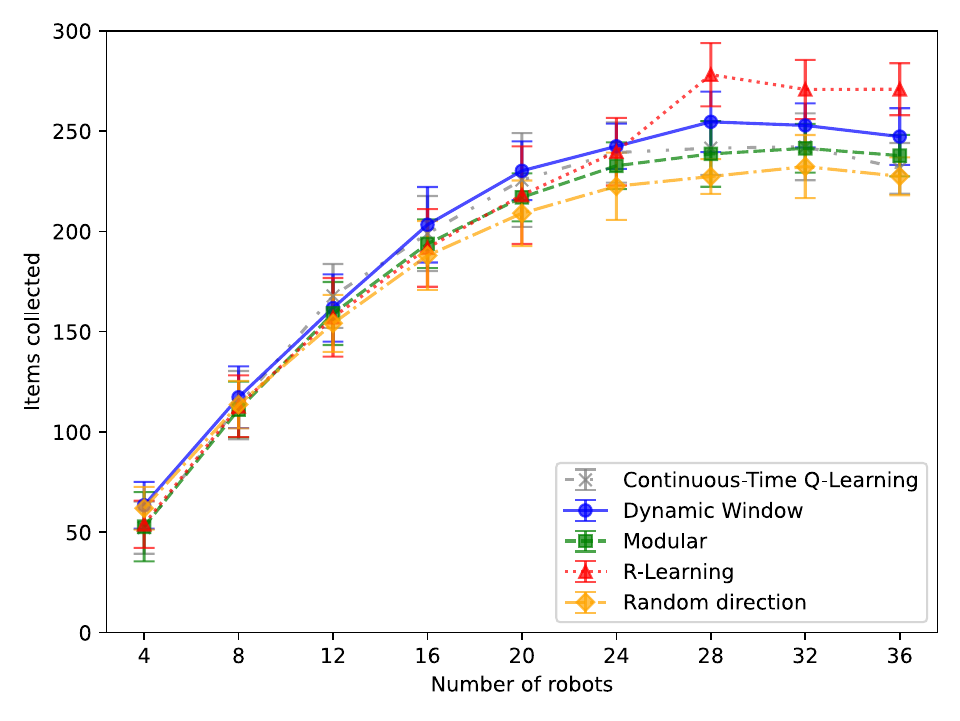}
    \caption{Results in Arena 3}
    \label{fig:mod3}
\end{figure}

Finally, in Figure~\ref{fig:mod3}, multiple bases make the density lower in comparison to a single base, resulting in fewer collisions as the robots can split between the bases. This makes this task easier, as evident by the higher puck collection values in comparison to Arena 1 (Fig.~\ref{fig:mod1}). Here, while the modular learning approach is consistently above the random selection baseline, it seems to face difficulties keeping up with dynamic window. We believe that the presence of two bases leads to ambiguous modular states, which the stochastic procedure of the council translates into inconsistent policies.

\subsection{Modular learning is robust to reward selection}\label{sec:robust}
As previously discussed, we chose to use the difference reward ($\diffr$) in all experiments, as it aligns individual and swarm goals. In practice, its use by robots involves a necessary approximation of others' overheads~\cite{royal25rational}. As it is given that such approximations can be surprisingly brittle, we wanted to test the robustness of the modular representation to changes in the reward value.

\begin{figure}[htbp]
    \includegraphics[width=0.8\linewidth]{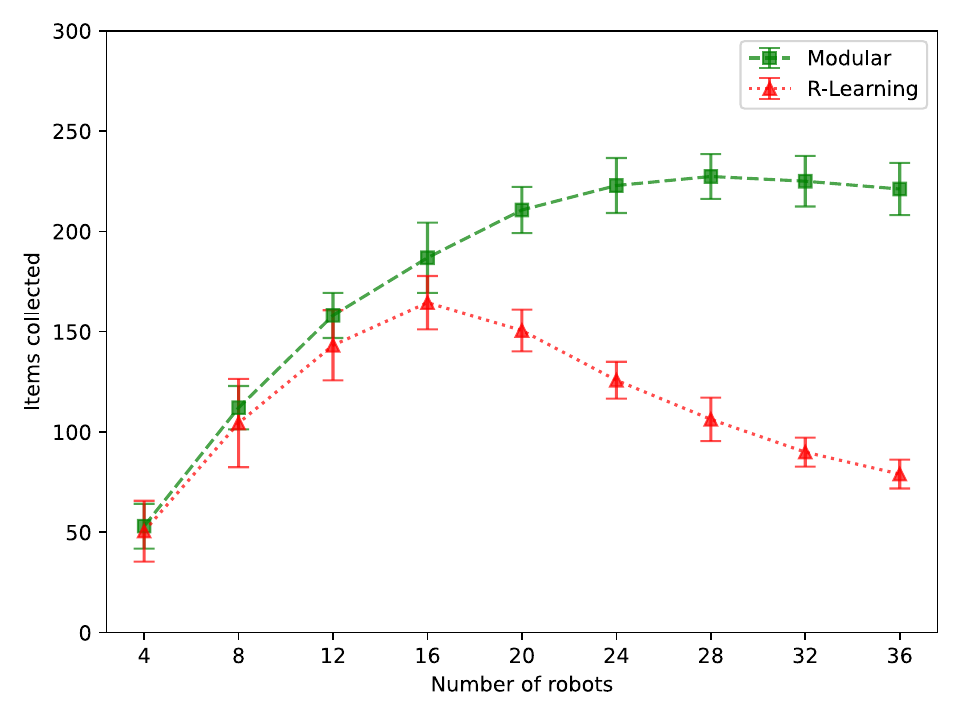}
    \caption{Modular and R-learning algorithms using $\diffr$ (Eq.~\ref{eq:diffr}) vs $\self$ (Eq.~\ref{eq:self}) in Arena 1.}\label{fig:robust1}
\end{figure}

\begin{figure}[htbp]
    \includegraphics[width=0.8\linewidth]{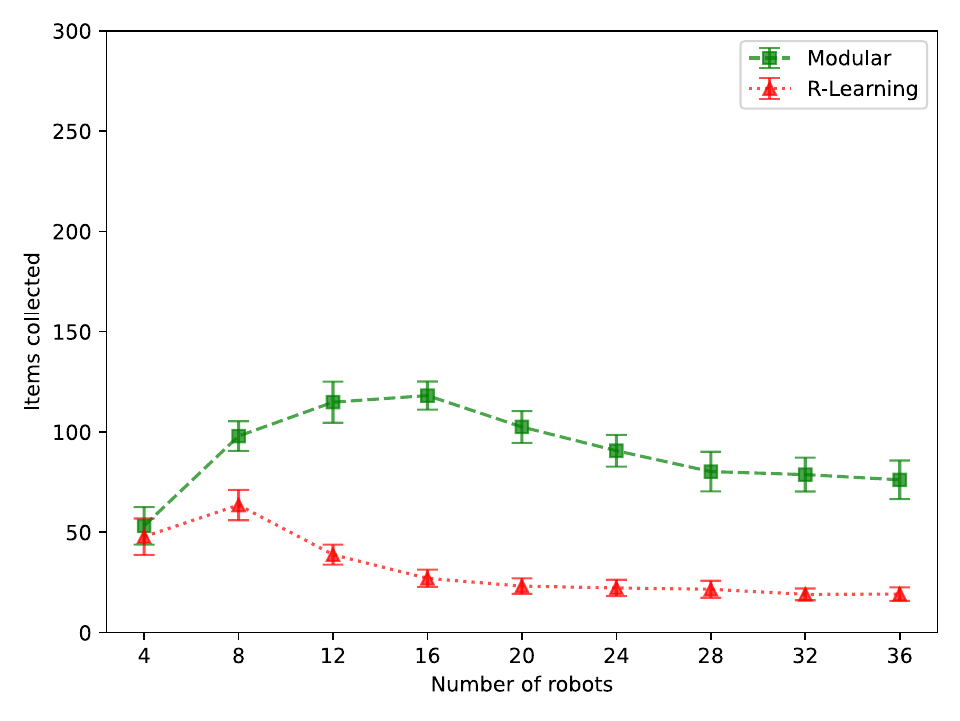}
    \caption{Modular and R-learning algorithms using $\diffr$ (Eq.~\ref{eq:diffr}) vs $\self$ (Eq.~\ref{eq:self}) in Arena 2.}\label{fig:robust2}
    \end{figure}

\begin{figure}[htbp]
    \includegraphics[width=0.8\linewidth]{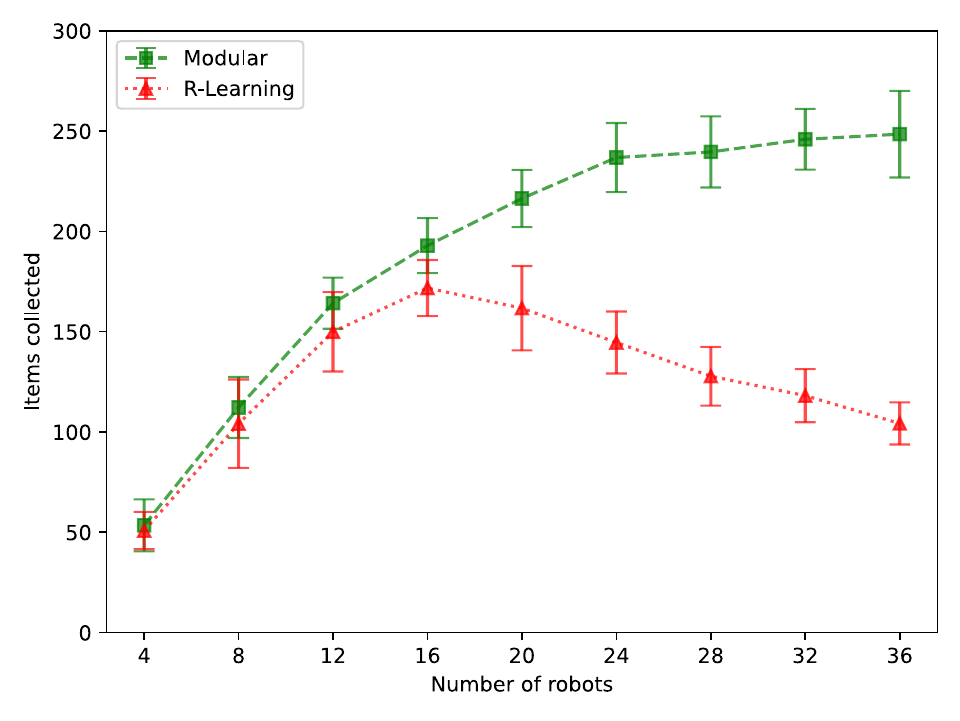}
    \caption{Modular and R-learning algorithms using $\diffr$ (Eq.~\ref{eq:diffr}) vs $\self$ (Eq.~\ref{eq:self}) in Arena 3.}\label{fig:robust3}
\end{figure}

We therefore test the modular representation and the R-learning algorithm performance, when the reward was changed from $\diffr$ (Eq.~\ref{eq:diffr}) to the raw self-interested reward $\self$ (Eq.~\ref{eq:self}), treating non-collision intervals as productive, and collision-handling intervals as costly. Figures~\ref{fig:robust1}--\ref{fig:robust3} show the results. The modular representation maintains its performance using the changed reward, while the stateful R-learning algorithm shows a dramatic change for the worse. As of now, we do not have an explanation for the robustness of the modular representation, and leave detailed investigation of this to future work.

\subsection{Vectorial vs Algorithmic Action Spaces}\label{sec:algspace}
As discussed, the modular method requires a vectorial action space to determine which action to choose, as collision-avoidance algorithms are atomic, so they cannot be merged by the aggregating mechanism of the council. This raises a problem when there is a disagreement between two or more decomposed learning processes. However, given that reliance on such algorithms as the action-space is so prominent in previous work~\cite{rybski98performance,Agression2,aij07avi,icra10dan,aamas19},
we wanted to evaluate whether the transition to the simpler vectorial action space leads to a loss in overall performance.

Since the modular representation cannot be used with algorithms, we instead used the R-learning algorithm described above, allowing it to select between three algorithms to respond to a collision state: \textit{Repel}~\cite{aij07avi}, \textit{dynamic window}~\cite{dynamic-window97} and aggression~\cite{Agression2}. 

\begin{figure}[htbp]
    \includegraphics[width=0.8\linewidth]{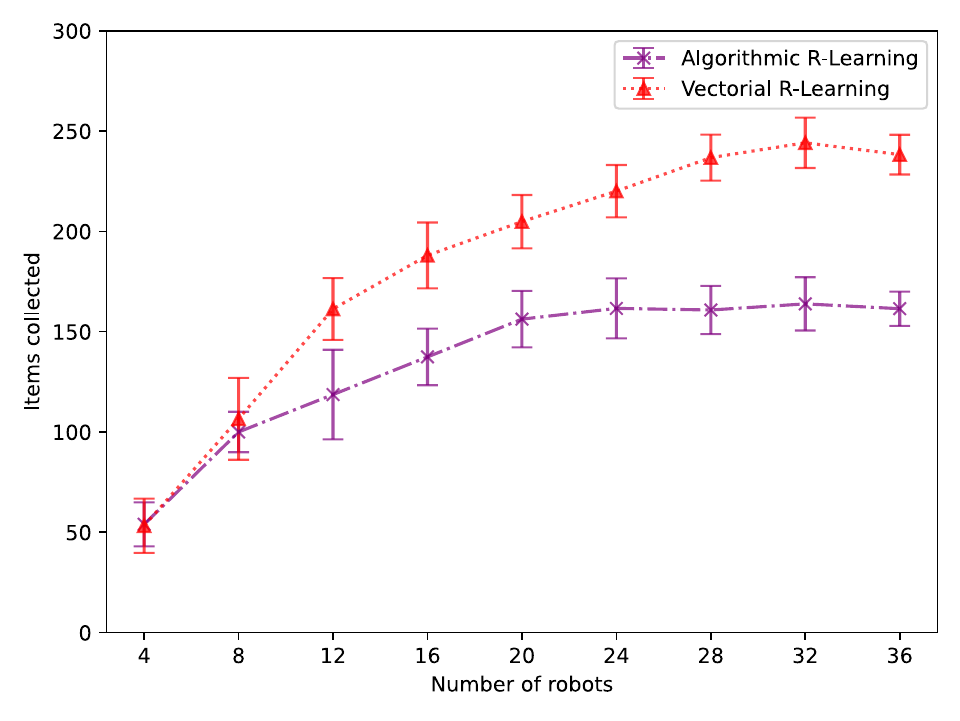}
    \caption{A comparison of learning in vectorial vs. algorithmic action spaces, in Arena 1}\label{fig:algspace1}
\end{figure}

\begin{figure}[htbp]
    \includegraphics[width=0.8\linewidth]{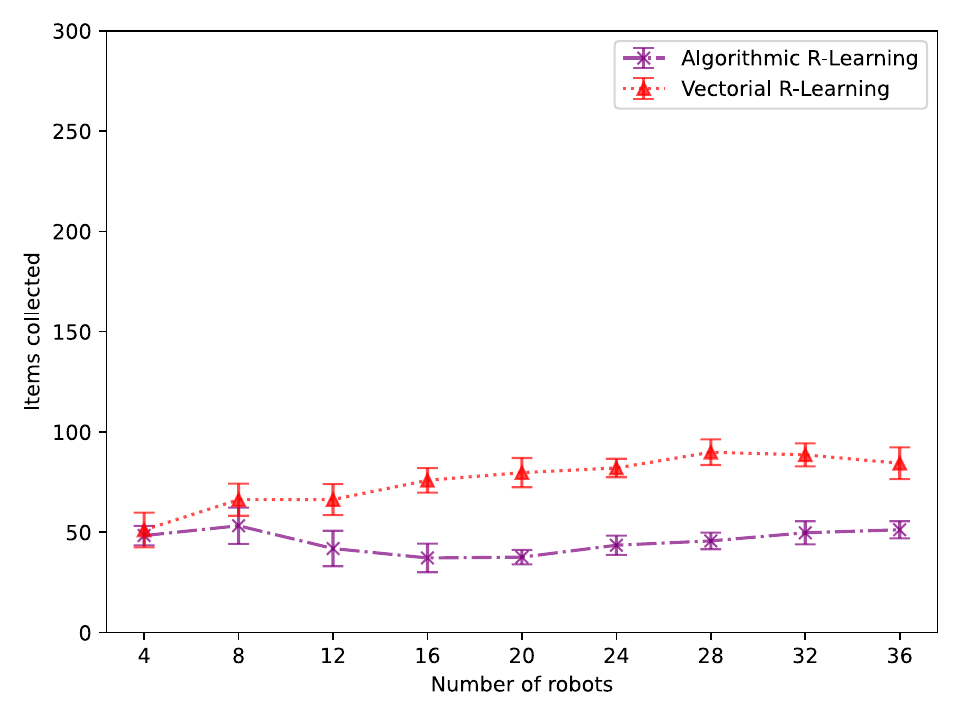}
    \caption{A comparison of learning in vectorial vs. algorithmic action spaces, in Arena  2}\label{fig:algspace2}
\end{figure}

\begin{figure}[htbp]
    \includegraphics[width=0.8\linewidth]{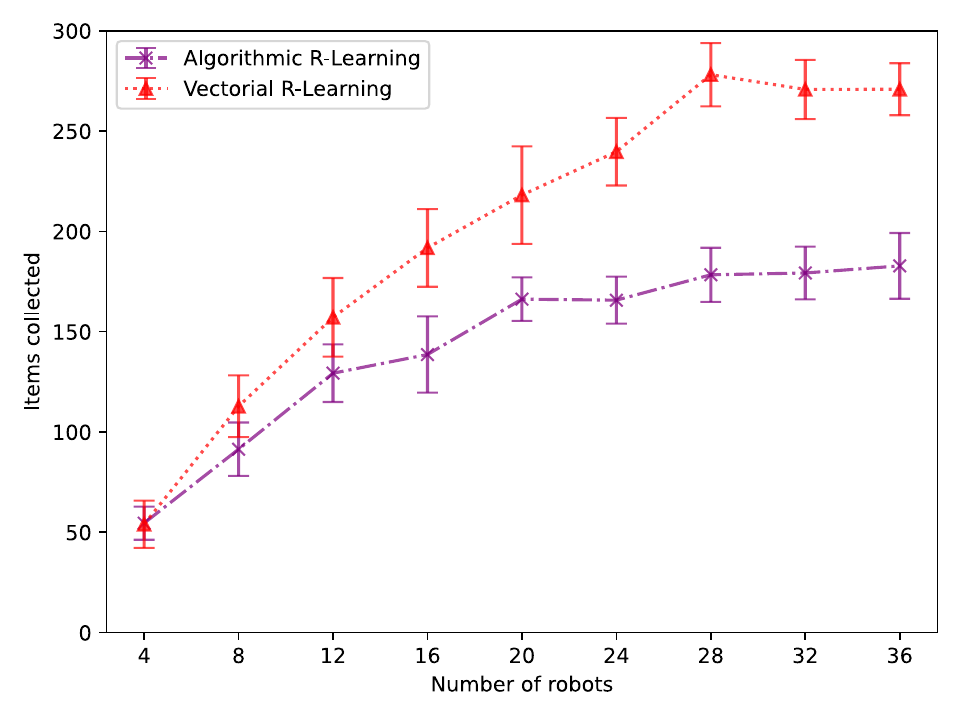}
    \caption{A comparison of learning in vectorial vs. algorithmic action spaces, in Arena  3}\label{fig:algspace3}
\end{figure}

The results appear in Figures~\ref{fig:algspace1}--\ref{fig:algspace3}. The figure shows that learning to use vectorial actions leads to improved results over learning to use collision-avoidance algorithms to handle collisions.

%% file: fin.tex
Cooperative robot swarms can learn how to interact effectively with others, but to do so, they need to overcome the hurdle of state explosion. This is not trivial for physical swarm robots, whose memory is often well below 200KB. We presented a modular state representation approach that takes advantage of the natural decomposition of spatial states and the natural composability of the vectorial action space: a separate learning process is assigned to each range and bearing sensor. This leads to a total number of states that is linear in the number of sensors, rather than exponential. The state-space of each independent learning process is restricted to the readings from the sensor, but its action space is over the entire robot (i.e., all potential headings). The selected actions from all sensors are combined by a mechanism that samples the action preference distribution to choose an action to be executed by the robot.

We have conducted extensive experiments, contrasting the performance of the modular representation with stateful reinforcement learners. The results demonstrate that the modular approach, while not always outperforming others, is robust and generally offers good performance, with dramatic memory savings. In addition, the modular representation can utilize rewards that cause the stateful learners to fail.

We believe modular representations are particularly well suited to spatial states, as they work in similar state spaces as potential fields, and naturally admit a composable action space (vectors of motion). Here, however, they are used as part of a learning process. We intend to explore the use of this modular representation in other swarm robotics tasks as well. In addition, we plan to investigate the use of learning at the aggregation level.